\documentclass[twoside]{article}

\usepackage[accepted]{aistats2026}
\usepackage[protrusion=false]{microtype}
\usepackage[T1]{fontenc}
\usepackage{bbding}
\usepackage[weather, misc]{ifsym}
\usepackage{tikz}
\usetikzlibrary{automata, positioning, shapes, shapes.multipart, arrows, arrows.meta, calc, backgrounds, overlay-beamer-styles, decorations.pathreplacing, 3d, patterns.meta, decorations.pathreplacing, shadows.blur, quotes, arrows.meta, positioning, trees}
\usepackage{multirow}
\usepackage{svg}
\usepackage{subcaption}
\usepackage[table, dvipsnames]{xcolor}
\usepackage{booktabs}
\usepackage{amsmath}
\usepackage{amssymb}
\usepackage{hyperref}

\newcommand{\attn}{\textsc{attn}}
\newcommand{\mlp}{\textsc{mlp}}
\newcommand{\other}{\textsc{others}}
\newcommand{\all}{{\attn ~+ \mlp ~+ \other}}
\newcommand{\attnmlponly}{{\attn ~+ \mlp}}
\newcommand{\attnonly}{{\attn}}

\DeclareUnicodeCharacter{2212}{-}
\usepackage{algorithm}
\usepackage{algpseudocode}

\newcommand{\nyud}{NYUD-v2}

\usepackage[round]{natbib}

\begin{document}

\runningtitle{Parameter-Efficient Multi-Task Learning via Progressive Task-Specific Adaptation}
\runningauthor{Gangwar, Rangi, Deshmukh, Rahmanian, Dattatreya, Kani}

\twocolumn[

\aistatstitle{Parameter-Efficient Multi-Task Learning via Progressive\\Task-Specific Adaptation}

\aistatsauthor{Neeraj Gangwar\textsuperscript{\normalfont \textdagger}
\quad
Anshuka Rangi\textsuperscript{\normalfont \S}
\quad
Rishabh Deshmukh\textsuperscript{\normalfont \S}
\quad
Holakou Rahmanian\textsuperscript{\normalfont \S}}

\aistatsauthor{Yesh Dattatreya\textsuperscript{\normalfont \S}
\quad
Nickvash Kani\textsuperscript{\normalfont \textdagger}\vspace{0.5em}}

\aistatsaddress{\textsuperscript{\textdagger}University of Illinois Urbana-Champaign
\quad
\textsuperscript{\S}Amazon\\
{\small \texttt{gangwar2@illinois.edu}, \texttt{anshuka@amazon.com}, \texttt{derishab@amazon.com}}}]

\begin{abstract}
\hyphenpenalty=3000
Parameter-efficient fine-tuning methods have emerged as a promising solution for adapting pre-trained models to various downstream tasks. While these methods perform well in single-task learning, extending them to multi-task learning exacerbates common issues, such as task interference and negative transfer, due to the limited number of trainable parameters. To address these challenges, we introduce progressive task-specific multi-task adaptation, a novel parameter-efficient approach for multi-task learning. Our approach introduces adapter modules that are shared in early layers and become increasingly task-specific in later layers. Additionally, we propose a gradient-based approach for computing task similarity and use this measure to allocate similar tasks to the shared adapter modules. To evaluate our approach, we adapt Swin and Pyramid Vision Transformers on PASCAL and \nyud{}. On both datasets, our approach outperforms prior parameter-efficient multi-task methods while using fewer trainable parameters.
\end{abstract}
    
\section{INTRODUCTION}
\label{sec:introduction}
\begin{figure}[t]
    \centering
    \subfloat[\label{fig:peft_polyhistor}]{
        \begin{tikzpicture}[font={\fontfamily{phv}\selectfont\tiny}, text=darkgray!90, draw=darkgray!90]
    \tikzset{
        shadowbox/.style={rounded corners=0.05cm, blur shadow={shadow blur steps=5}}
    }
    \node [shadowbox, draw=OliveGreen, fill=OliveGreen!15, minimum width=0.75cm, minimum height=1.5cm] (B1) {\SnowflakeChevron};

    \node [shadowbox, draw=OliveGreen, fill=OliveGreen!15, minimum width=0.75cm, minimum height=1.5cm, right=of B1, right=1cm] (B2) {\SnowflakeChevron};

    \node [shadowbox, draw=OliveGreen, fill=OliveGreen!15, minimum width=0.75cm, minimum height=1.5cm, right=of B2, right=1cm] (B3) {\SnowflakeChevron};

    \node [shadowbox, draw=BurntOrange, fill=BurntOrange!20, minimum width=0.6cm, minimum height=0.3cm, right=of B3] (D1) {};

    \node [shadowbox, draw=CadetBlue, fill=CadetBlue!20, minimum width=0.6cm, minimum height=0.3cm, above=of D1, above=0.2cm] (D0) {};

    \node [shadowbox, draw=CarnationPink, fill=CarnationPink!20, minimum width=0.6cm, minimum height=0.3cm, below=of D1, below=0.2cm] (D2) {};

    \node [shadowbox, draw=BurntOrange, fill=BurntOrange!20, minimum width=0.3cm, minimum height=0.3cm] (T11) at ($(B1)!0.5!(B2)$) {};

    \node [shadowbox, draw=CadetBlue, fill=CadetBlue!20, minimum width=0.3cm, minimum height=0.3cm, above=of T11, above=0.2cm] (T10) {};

    \node [shadowbox, draw=CarnationPink, fill=CarnationPink!20, minimum width=0.3cm, minimum height=0.3cm, below=of T11, below=0.2cm] (T12) {};

    \node [shadowbox, draw=BurntOrange, fill=BurntOrange!20, minimum width=0.3cm, minimum height=0.3cm] (T21) at ($(B2)!0.5!(B3)$) {};

    \node [shadowbox, draw=CadetBlue, fill=CadetBlue!20, minimum width=0.3cm, minimum height=0.3cm, above=of T21, above=0.2cm] (T20) {};

    \node [shadowbox, draw=CarnationPink, fill=CarnationPink!20, minimum width=0.3cm, minimum height=0.3cm, below=of T21, below=0.2cm] (T22) {};

    \node [shadowbox, draw=BurntOrange, fill=BurntOrange!20, minimum width=0.3cm, minimum height=0.3cm] (T31) at ($(B3)!0.5!(D1)$) {};

    \node [shadowbox, draw=CadetBlue, fill=CadetBlue!20, minimum width=0.3cm, minimum height=0.3cm, above=of T31, above=0.2cm] (T30) {};

    \node [shadowbox, draw=CarnationPink, fill=CarnationPink!20, minimum width=0.3cm, minimum height=0.3cm, below=of T31, below=0.2cm] (T32) {};

    \node [right=of D0, right=0.5cm] (T0) {Task 1};
    \node [right=of D1, right=0.5cm] (T1) {Task 2};
    \node [right=of D2, right=0.5cm] (T2) {Task 3};

    \node [below=of D2, below=0.5cm] (DEC_DESC) {Decoders};
    \node [below=of T22, below=0.4cm] (ADA_DESC) {Individual single-task adaptation};
    \node [above=of B2, above=0.3cm] (BCK_DESC) {Backbone};

    \draw [-Stealth] (D0) -- (T0);
    \draw [-Stealth] (D1) -- (T1);
    \draw [-Stealth] (D2) -- (T2);

    \draw [-Stealth] (B1) -- (T10);
    \draw [-Stealth] (B1) -- (T11);
    \draw [-Stealth] (B1) -- (T12);

    \draw [-Stealth] (T10) -- (B2);
    \draw [-Stealth] (T11) -- (B2);
    \draw [-Stealth] (T12) -- (B2);

    \draw [-Stealth] (B2) -- (T20);
    \draw [-Stealth] (B2) -- (T21);
    \draw [-Stealth] (B2) -- (T22);

    \draw [-Stealth] (T20) -- (B3);
    \draw [-Stealth] (T21) -- (B3);
    \draw [-Stealth] (T22) -- (B3);

    \draw [-Stealth] (B3) -- (T30);
    \draw [-Stealth] (B3) -- (T31);
    \draw [-Stealth] (B3) -- (T32);

    \draw [-Stealth] (T30) -- (D0);
    \draw [-Stealth] (T31) -- (D1);
    \draw [-Stealth] (T32) -- (D2);

    \draw [latex-, densely dashed] ($(D2.south) - (0, 0.1cm)$) -- (DEC_DESC);
    \draw [latex-, densely dashed] ($(T12.south) - (0, 0.1cm)$) -- (ADA_DESC);
    \draw [latex-, densely dashed] ($(T22.south) - (0, 0.1cm)$) -- (ADA_DESC);
    \draw [latex-, densely dashed] ($(T32.south) - (0, 0.1cm)$) -- (ADA_DESC);

    \draw [latex-, densely dashed] ($(B1.north) + (0, 0.1cm)$) -- (BCK_DESC);
    \draw [latex-, densely dashed] ($(B2.north) + (0, 0.1cm)$) -- (BCK_DESC);
    \draw [latex-, densely dashed] ($(B3.north) + (0, 0.1cm)$) -- (BCK_DESC);

\end{tikzpicture}
    }

    \subfloat[\label{fig:peft_mtlora}]{
        \begin{tikzpicture}[font={\fontfamily{phv}\selectfont\tiny}, text=darkgray!90, draw=darkgray!90]
    \tikzset{
        roundbox/.style={rounded corners=0.05cm},
        shadowbox/.style={roundbox, blur shadow={shadow blur steps=8}}
    }
    \node [shadowbox, draw=OliveGreen, fill=OliveGreen!15, minimum width=0.75cm, minimum height=1.5cm] (B1) {\SnowflakeChevron};

    \node [shadowbox, draw=OliveGreen, fill=OliveGreen!15, minimum width=0.75cm, minimum height=1.5cm, right=of B1, right=1cm] (B2) {\SnowflakeChevron};

    \node [shadowbox, draw=OliveGreen, fill=OliveGreen!15, minimum width=0.75cm, minimum height=1.5cm, right=of B2, right=1cm] (B3) {\SnowflakeChevron};

    \node [shadowbox, draw=BurntOrange, fill=BurntOrange!20, minimum width=0.6cm, minimum height=0.3cm, right=of B3] (D1) {};

    \node [shadowbox, draw=CadetBlue, fill=CadetBlue!20, minimum width=0.6cm, minimum height=0.3cm, above=of D1, above=0.2cm] (D0) {};

    \node [shadowbox, draw=CarnationPink, fill=CarnationPink!20, minimum width=0.6cm, minimum height=0.3cm, below=of D1, below=0.2cm] (D2) {};

    \node [shadowbox, draw=Gray, fill=Gray!20, minimum width=0.5cm, minimum height=1.4cm] (TB1) at ($(B1)!0.5!(B2)$) {};

    \node [roundbox, draw=BurntOrange, fill=BurntOrange!20, minimum width=0.3cm, minimum height=0.3cm] (T11) at ($(B1)!0.5!(B2)$) {};

    \node [roundbox, draw=CadetBlue, fill=CadetBlue!20, minimum width=0.3cm, minimum height=0.3cm, above=of T11, above=0.1cm] (T10) {};

    \node [roundbox, draw=CarnationPink, fill=CarnationPink!20, minimum width=0.3cm, minimum height=0.3cm, below=of T11, below=0.1cm] (T12) {};

    \node [shadowbox, draw=Gray, fill=Gray!20, minimum width=0.5cm, minimum height=1.4cm] (TB2) at ($(B2)!0.5!(B3)$) {};

    \node [roundbox, draw=BurntOrange, fill=BurntOrange!20, minimum width=0.3cm, minimum height=0.3cm] (T21) at ($(B2)!0.5!(B3)$) {};

    \node [roundbox, draw=CadetBlue, fill=CadetBlue!20, minimum width=0.3cm, minimum height=0.3cm, above=of T21, above=0.1cm] (T20) {};

    \node [roundbox, draw=CarnationPink, fill=CarnationPink!20, minimum width=0.3cm, minimum height=0.3cm, below=of T21, below=0.1cm] (T22) {};

    \node [shadowbox, draw=Gray, fill=Gray!20, minimum width=0.5cm, minimum height=1.4cm] (TB3) at ($(B3)!0.5!(D1)$) {};

    \node [roundbox, draw=BurntOrange, fill=BurntOrange!20, minimum width=0.3cm, minimum height=0.3cm] (T31) at ($(B3)!0.5!(D1)$) {};

    \node [roundbox, draw=CadetBlue, fill=CadetBlue!20, minimum width=0.3cm, minimum height=0.3cm, above=of T31, above=0.1cm] (T30) {};

    \node [roundbox, draw=CarnationPink, fill=CarnationPink!20, minimum width=0.3cm, minimum height=0.3cm, below=of T31, below=0.1cm] (T32) {};

    \node [right=of D0, right=0.5cm] (T0) {Task 1};
    \node [right=of D1, right=0.5cm] (T1) {Task 2};
    \node [right=of D2, right=0.5cm] (T2) {Task 3};

    \node [below=of D2, below=0.5cm] (DEC_DESC) {Decoders};
    \node [below=of T22, below=0.5cm] (ADA_DESC) {Shared multi-task adaptation};
    \node [above=of B2, above=0.3cm] (BCK_DESC) {Backbone};

    \draw [-Stealth] (D0) -- (T0);
    \draw [-Stealth] (D1) -- (T1);
    \draw [-Stealth] (D2) -- (T2);

    \draw [-Stealth] (B1) -- (TB1);
    \draw [-Stealth] (TB1) -- (B2);

    \draw [-Stealth] (B2) -- (TB2);
    \draw [-Stealth] (TB2) -- (B3);

    \draw [-Stealth] (B3) -- (TB3);

    \draw [-Stealth] (TB3) -- (D0);
    \draw [-Stealth] (TB3) -- (D1);
    \draw [-Stealth] (TB3) -- (D2);

    \draw [latex-, densely dashed] ($(D2.south) - (0, 0.1cm)$) -- (DEC_DESC);
    \draw [latex-, densely dashed] ($(TB1.south) - (0, 0.1cm)$) -- (ADA_DESC);
    \draw [latex-, densely dashed] ($(TB2.south) - (0, 0.1cm)$) -- (ADA_DESC);
    \draw [latex-, densely dashed] ($(TB3.south) - (0, 0.1cm)$) -- (ADA_DESC);

    \draw [latex-, densely dashed] ($(B1.north) + (0, 0.1cm)$) -- (BCK_DESC);
    \draw [latex-, densely dashed] ($(B2.north) + (0, 0.1cm)$) -- (BCK_DESC);
    \draw [latex-, densely dashed] ($(B3.north) + (0, 0.1cm)$) -- (BCK_DESC);

\end{tikzpicture}
    }

    \subfloat[\label{fig:peft_tglora}]{
        \begin{tikzpicture}[font={\fontfamily{phv}\selectfont\tiny}, text=darkgray!90, draw=darkgray!90]
    \tikzset{
        roundbox/.style={rounded corners=0.05cm},
        shadowbox/.style={roundbox, blur shadow={shadow blur steps=8}}
    }
    \node [shadowbox, draw=OliveGreen, fill=OliveGreen!15, minimum width=0.75cm, minimum height=1.5cm] (B1) {\SnowflakeChevron};

    \node [shadowbox, draw=OliveGreen, fill=OliveGreen!15, minimum width=0.75cm, minimum height=1.5cm, right=of B1, right=1cm] (B2) {\SnowflakeChevron};

    \node [shadowbox, draw=OliveGreen, fill=OliveGreen!15, minimum width=0.75cm, minimum height=1.5cm, right=of B2, right=1cm] (B3) {\SnowflakeChevron};

    \node [shadowbox, draw=BurntOrange, fill=BurntOrange!20, minimum width=0.6cm, minimum height=0.3cm, right=of B3] (D1) {};

    \node [shadowbox, draw=CadetBlue, fill=CadetBlue!20, minimum width=0.6cm, minimum height=0.3cm, above=of D1, above=0.2cm] (D0) {};

    \node [shadowbox, draw=CarnationPink, fill=CarnationPink!20, minimum width=0.6cm, minimum height=0.3cm, below=of D1, below=0.2cm] (D2) {};

    \node [shadowbox, draw=Gray, fill=Gray!20, minimum width=0.5cm, minimum height=1.4cm] (TB10) at ($(B1)!0.5!(B2)$) {};

    \node [roundbox, draw=BurntOrange, fill=BurntOrange!20, minimum width=0.3cm, minimum height=0.3cm] (T11) at ($(B1)!0.5!(B2)$) {};

    \node [roundbox, draw=CadetBlue, fill=CadetBlue!20, minimum width=0.3cm, minimum height=0.3cm, above=of T11, above=0.1cm] (T10) {};

    \node [roundbox, draw=CarnationPink, fill=CarnationPink!20, minimum width=0.3cm, minimum height=0.3cm, below=of T11, below=0.1cm] (T12) {};

    \node [shadowbox, draw=Gray, fill=Gray!20, minimum width=0.5cm, minimum height=1cm, yshift=0.3cm] (TB20) at ($(B2)!0.5!(B3)$) {};

    \node [shadowbox, draw=Gray, fill=Gray!20, minimum width=0.5cm, minimum height=0.5cm, yshift=-0.6cm] (TB21) at ($(B2)!0.5!(B3)$) {};

    \node [roundbox, draw=BurntOrange, fill=BurntOrange!20, minimum width=0.3cm, minimum height=0.3cm, yshift=0.1cm] (T21) at ($(B2)!0.5!(B3)$) {};

    \node [roundbox, draw=CadetBlue, fill=CadetBlue!20, minimum width=0.3cm, minimum height=0.3cm, above=of T21, above=0.1cm] (T20) {};

    \node [roundbox, draw=CarnationPink, fill=CarnationPink!20, minimum width=0.3cm, minimum height=0.3cm, below=of TB20, below=0.23cm] (T22) {};


    \node [shadowbox, draw=BurntOrange, fill=BurntOrange!20, minimum width=0.3cm, minimum height=0.3cm] (T31) at ($(B3)!0.5!(D1)$) {};

    \node [shadowbox, draw=CadetBlue, fill=CadetBlue!20, minimum width=0.3cm, minimum height=0.3cm, above=of T31, above=0.2cm] (T30) {};

    \node [shadowbox, draw=CarnationPink, fill=CarnationPink!20, minimum width=0.3cm, minimum height=0.3cm, below=of T31, below=0.2cm] (T32) {};

    \node [right=of D0, right=0.5cm] (T0) {Task 1};
    \node [right=of D1, right=0.5cm] (T1) {Task 2};
    \node [right=of D2, right=0.5cm] (T2) {Task 3};

    \node [below=of D2, below=0.5cm] (DEC_DESC) {Decoders};
    \node [below=of T22, below=0.4cm, xshift=-0.2cm] (ADA_DESC) {Progressive task-specific multi-task adaptation};
    \node [above=of B2, above=0.3cm] (BCK_DESC) {Backbone};

    \draw [-Stealth] (D0) -- (T0);
    \draw [-Stealth] (D1) -- (T1);
    \draw [-Stealth] (D2) -- (T2);

    \draw [-Stealth] (B1) -- (TB1);
    \draw [-Stealth] (TB1) -- (B2);

    \draw [-Stealth] (B2) -- (TB20);
    \draw [-Stealth] (B2) -- (TB21);
    \draw [-Stealth] (TB20) -- (B3);
    \draw [-Stealth] (TB21) -- (B3);

    \draw [-Stealth] (B3) -- (T30);
    \draw [-Stealth] (B3) -- (T31);
    \draw [-Stealth] (B3) -- (T32);

    \draw [-Stealth] (T30) -- (D0);
    \draw [-Stealth] (T31) -- (D1);
    \draw [-Stealth] (T32) -- (D2);

    \draw [latex-, densely dashed] ($(D2.south) - (0, 0.1cm)$) -- (DEC_DESC);
    \draw [latex-, densely dashed] ($(TB1.south) - (0, 0.1cm)$) -- (ADA_DESC);
    \draw [latex-, densely dashed] ($(TB2.south) - (0, 0.2cm)$) -- (ADA_DESC);
    \draw [latex-, densely dashed] ($(TB3.south) - (0, 0.1cm)$) -- (ADA_DESC);

    \draw [latex-, densely dashed] ($(B1.north) + (0, 0.1cm)$) -- (BCK_DESC);
    \draw [latex-, densely dashed] ($(B2.north) + (0, 0.1cm)$) -- (BCK_DESC);
    \draw [latex-, densely dashed] ($(B3.north) + (0, 0.1cm)$) -- (BCK_DESC);

\end{tikzpicture}
    }

    \caption{(a) Single-task adaptation uses separate adapters per task, preventing knowledge transfer and increasing inference cost with more tasks. (b) Shared multi-task adaptation uses common adapters, reducing inference cost at the expense of potential task interference. (c) \textit{Progressive task-specific multi-task adaptation}, where adapters become increasingly task specific near the decoders, enables both knowledge transfer and task specialization.}
\end{figure}
Large language and vision models pre-trained on extensive datasets have demonstrated an unprecedented ability to understand and generate human-like text and images, perform complex reasoning, and analyze data \citep[e.g.,][]{achiam2023gpt, team2023gemini, gao2023llama, touvron2023llama, bai2024sequential}. These models can learn from a few demonstrations through in-context learning or can be adapted to various downstream tasks through targeted fine-tuning. Although fine-tuning may result in better performance compared to in-context learning, updating all model parameters demands substantial computational resources. Moreover, a separate copy of the model must be stored for each task, and this challenge only worsens as model sizes increase.

Several parameter-efficient methods have been proposed to address these challenges \citep[e.g.,][]{hu2022lora, li2021prefix, lester2021power, houlsby2019parameter, sung2021training, zhang2023adalora, liu2024dora}. These approaches either fine-tune a subset of the model's existing parameters or add new layers, training only these layers. In both cases, the number of trainable parameters remains small relative to the overall model size. These methods have achieved performance levels that nearly match those of full fine-tuning, offering a more favorable trade-off between trainable parameter count and downstream task performance. They also reduce storage requirements, as only the updated parameters must be stored separately for each task. These methods have been widely used for adapting large pre-trained models to various natural language processing (NLP) and vision tasks \citep[e.g.,][]{hu2023llm,chen2022adaptformer,jia2022visual,chen2023vision,xin2024parameter}. While they have proven highly effective for adapting single-task models, their applicability to multi-task learning (MTL) remains limited. One reason is the unique challenges associated with MTL. Due to parameter sharing across tasks, MTL can suffer from conflicting and dominating gradients. This results in interference and negative transfer between tasks, leading to significantly degraded performance \citep{yu2020gradient}. These problems become more pronounced in parameter-efficient multi-task learning (PEMTL), as the limited number of trainable parameters restricts the model's capacity to effectively separate task-specific knowledge. Although PEMTL is relatively underexplored, several works have made notable advancements \citep[e.g.,][]{pfeiffer2021adapterfusion, wang2022adamix, mahabadi2021parameter, liu2022polyhistor, agiza2024mtlora, baek2025tadformer}. Existing methods for adapting pre-trained models to perform multiple tasks fall into two main categories: (1) individual single-task adaptation and (2) shared multi-task adaptation.

\begin{figure}[t]
    \centering
    \resizebox{0.9\linewidth}{!}{
        \fontfamily{phv}\selectfont\large
        \includesvg[]{img/peft_params_vs_perf.svg}
    }
    \caption{Comparison of our proposed approach, progressive task-specific multi-task adaptation, with methods that use individual single-task and shared multi-task adaptations. These experiments are performed on the PASCAL dataset with Swin-Tiny. The numbers for individual single-task and shared multi-task adaptations are taken from \citet{agiza2024mtlora}.}
    \label{fig:peft_params_vs_perf}
\end{figure}

In individual single-task adaptation (Figure~\ref{fig:peft_polyhistor}), separate adapter modules\footnote{We use the term ``adapter module'' to refer to the parameters added to a model to adapt it for a downstream task.} are added for each downstream task, making it particularly effective when tasks are not related or have specific requirements. However, there are two major drawbacks. First, this method does not allow knowledge transfer among tasks, which may lead to suboptimal performance compared to models with shared parameters. Second, the computational cost of inference increases linearly with the number of tasks, as the adapter modules for each task must be executed individually. In shared multi-task adaptation (Figure~\ref{fig:peft_mtlora}), the adapter modules are shared across all tasks. This setup enables knowledge transfer among tasks and reduces inference cost compared to individual single-task adaptation. However, due to gradient conflicts, shared adapter modules can suffer from task interference and negative transfer, limiting task-specific learning.

To address these issues, we introduce \textit{progressive task-specific multi-task adaptation}, a parameter-efficient approach, illustrated in Figure~\ref{fig:peft_tglora}, that balances the trade-off between individual single-task and shared multi-task adaptations. In our approach, adapter modules are shared across all tasks in the initial layers and become increasingly task-specific toward the decoders. Structuring adapter modules in this manner effectively forms a branched multi-task network. Branched networks have been effective in full fine-tuning settings \citep{Lu2016FullyAdaptiveFS, Vandenhende2019BranchedMN, Brggemann2020AutomatedSF, Guo2020LearningTB}; however, branching significantly increases trainable parameters and storage requirements. Adapter modules offer a more efficient way of implementing branching while maintaining strong multi-task performance with substantially fewer trainable parameters. Our approach achieves an inference cost that falls between those of individual single-task and shared multi-task adaptations. To reduce conflicts among tasks, we assign \textit{similar} tasks to the shared adapter modules. We use a gradient-based method to compute task similarity, which introduces minimal computational overhead. We hypothesize that a progressive task-specific architecture could better handle task conflicts and improve performance with fewer trainable parameters. We compare our approach with existing PEMTL methods on the PASCAL and \nyud{} datasets. Note that our work complements methods that use gradient surgery to reduce gradient conflicts among tasks and can therefore be combined with any such method. In our evaluation, however, we assess our approach without incorporating such methods and employ standard fine-tuning.

\paragraph{Contributions.}
Our contributions are as follows:

\vspace{-0.5em}
\begin{enumerate}
    \item We introduce progressive task-specific adaptation, a parameter-efficient method for multi-task learning that bridges the gap between individual single-task and shared multi-task adaptations. It facilitates knowledge transfer among tasks and reduces interference through intelligent task grouping.

    \item We propose a gradient-based task similarity approach to assign similar tasks to the shared adapter modules. Our experiments show that grouping less similar tasks, identified by this approach, adversely impacts the multi-task model's performance, proving its effectiveness.

    \item We create a LoRA-based \citep{hu2022lora} layer named TGLoRA, which is used to implement the progressive task-specific architecture.

    \item Our approach achieves a better relative improvement over single-task fine-tuning and uses fewer trainable parameters than the current state-of-the-art PEMTL approaches. Figure~\ref{fig:peft_params_vs_perf} shows a comparison of our approach with existing approaches on the PASCAL dataset.
\end{enumerate}

\vspace{-0.5em}

Our code is publicly available on GitHub.\footnote{\href{https://github.com/neerajgangwar/progressive-task-specific-adaptation}{\texttt{https://github.com/neerajgangwar/progressive\\-task-specific-adaptation}}}

\section{RELATED WORKS}
\label{sec:related_works}
\paragraph{Parameter-Efficient Fine-Tuning.}
To achieve a better trade-off between the number of trainable parameters and performance on downstream tasks, various methods have been proposed. These approaches fine-tune a small set of parameters while the rest of the model is frozen. One set of approaches adds new layers to the model and fine-tunes only these. For example, \citet{houlsby2019parameter} and \citet{pfeiffer2020adapterhub} place Adapter layers after the attention and MLP layers and fine-tune these newly added layers to achieve a comparable performance to full fine-tuning on NLP datasets. \citet{hu2022lora} propose LoRA, a method to efficiently fine-tune large language models by adding low-rank decomposition matrices to their weights, significantly reducing the number of trainable parameters. Several subsequent works have proposed variations of LoRA, for example, AdaLoRA \citep{zhang2023adalora}, LoRA Dropout \citep{lin2024lora}, DoRA \citep{liu2024dora}, and LoRA+ \citep{hayou2024lora+}, among others. Another line of research adds trainable prompts to the input and fine-tunes only these while keeping the model frozen \citep{li2021prefix, lester2021power}. Methods like BitFit \citep{zaken2022bitfit} and FISH \citep{sung2021training} fine-tune a small subset of the model's parameters without adding any new parameters. While parameter-efficient methods have been widely applied to NLP tasks, several works have explored their application for vision tasks. \citet{jia2022visual} propose visual prompt tuning for adapting Vision Transformers. \citet{gao2023llama} propose Llama-Adapter V2, which leverages both prefix-tuning and adapter techniques.

\paragraph{Multi-Task Learning.}
MTL in computer vision aims to train a model to perform multiple related tasks simultaneously, such as object detection, segmentation, depth estimation, etc. By sharing representations across tasks, MTL allows models to leverage shared information, potentially improving performance and efficiency compared to training separate models for each task \citep{crawshaw2020multi, vandenhende2021multi, yu2024unleashing}. However, MTL faces common challenges, such as task interference, negative transfer, and task dominance \citep{yu2020gradient}. Several methods have been proposed to mitigate these issues in a full fine-tuning setting \citep{chen2018gradnorm, kendall2018multi, maninis2019attentive, senushkin2023independent, ban2024fair}. Another line of research has focused on branched multi-task networks \citep{Lu2016FullyAdaptiveFS, Vandenhende2019BranchedMN, Brggemann2020AutomatedSF, Guo2020LearningTB}. The branching aims to balance shared representations and task-specific learning to improve performance and efficiency. While some works use neural architecture search \citep{Brggemann2020AutomatedSF} or reinforcement learning \citep{Guo2020LearningTB}, our approach aligns with methods that leverage task affinity to group similar tasks \citep{Vandenhende2019BranchedMN, fifty2021efficiently}. However, unlike these methods, our approach does not train separate models to compute task similarity, making it more efficient. Other methods to compute task similarity have also used Fisher information \citep{achille2019task2vec}. Instead, we compute similarity directly from gradients.

\paragraph{Parameter-Efficient MTL.}
Although PEMTL remains relatively underexplored, several works have made notable contributions. To address the lack of knowledge transfer in individual single-task adaptation, Hyperformer \citep{mahabadi2021parameter} uses a common hypernetwork to generate Adapter module parameters for different tasks, training only the hypernetwork. This approach facilitates knowledge transfer across tasks via the shared hypernetwork. \citet{yang2025mtl} propose MTL-LoRA for NLP tasks, which uses a shared down-projection matrix, task-specific transformation matrices, and multiple up-projection matrices. Polyhistor \citep{liu2022polyhistor} is a parameter-efficient version of Hyperformer for dense vision tasks. While this approach enables knowledge transfer, the inference cost still increases linearly with the number of tasks. To address this, \citet{agiza2024mtlora} introduce MTLoRA, closely approximating shared multi-task adaptation. Unlike shared multi-task adaptation, which relies solely on shared modules, MTLoRA incorporates task-specific modules in certain layers to enable task-specific learning. Although this mitigates some interference, most layers in MTLoRA remain shared across tasks, leading to task conflicts, and the number of task-specific modules scales linearly with the number of tasks. TADFormer \citep{baek2025tadformer} uses task-aware feature adaptation for dense prediction tasks. Additionally, mixture-of-experts has also been used to implement PEMTL \citep[e.g.,][]{yang2024multi, lin2024pemt}.

\section{PROPOSED APPROACH}
\label{sec:proposed_approach}
In our approach, adapter modules are shared across tasks in early layers and gradually become more task-specific in later layers. To reduce task interference, we assign similar tasks to an adapter module. This intelligent parameter sharing enables effective knowledge transfer among tasks while reducing task conflicts, addressing both the knowledge transfer limitations of individual single-task adaptation and the task interference challenges of shared multi-task adaptation. This architecture achieves better inference cost than individual single-task adaptation by significantly reducing the number of dedicated adapter modules. With a moderately higher inference cost than shared multi-task adaptation, this architecture offers significant performance improvement.

We define a \textit{task group} as a set of tasks assigned to a single adapter module. Adding multiple modules of this type within a layer effectively creates copies of that layer, with each copy specialized for a task group. Each copy's parameters are updated based on the combined loss for its respective task group. As a result, the progressive task-specific adaptation shown in Figure~\ref{fig:peft_tglora} is equivalent to transforming a linear architecture into a branched structure, where the task-specific decoders serve as the leaves. The tree-like structure ensures the task specificity of the adapter modules. Notably, the adapter modules in the final layer can be shared across multiple tasks rather than being task-specific. This helps limit the proliferation of adapter modules and manage the resulting tree's width in the case of extreme MTL, addressing the linear expansion of task-specific modules observed in MTLoRA.

The number of task groups in a layer is determined by a predefined computational budget, such as parameter count or floating-point operations per second (FLOPS). To maintain the task-specificity of adapter modules, two tasks in a layer's task group must not belong to different task groups in any preceding layers.

\subsection{Task Similarity}
\label{sec:task_similarity}
We use the notion of gradient conflicts from the MTL literature to compute the similarity between a pair of tasks. Formally, consider a set of tasks $\mathcal{T} = \left\lbrace 1, 2, \dots T \right\rbrace$ and model parameters $\left\lbrace \theta_s \right\rbrace \cup \left\lbrace \theta_t \vert t \in \mathcal{T} \right\rbrace$, where $\theta_s$ and $\theta_t$ represent the shared and task-specific parameters, respectively. Let $\mathcal{D}_t = \left\lbrace x_{t, i} \;\vert\; 1 \le i \le \vert \mathcal{D}_t \vert \right\rbrace$ represent the training dataset for task $t$. We define the similarity between two tasks based on the gradient of each task’s loss with respect to the shared parameters. Specifically, the similarity between tasks $t$ and $t^\prime$ is given by
\begin{equation}
    \label{eq:task_sim}
    \begin{gathered}
    \mathbb{E}_{x \sim \mathcal{D}_t, x^{\prime} \sim \mathcal{D}_{t^{\prime}}} \left[ S \left( g \left(x, t \right), g \left( x^\prime, t^\prime \right) \right) \right] \\
    g(x, t) \triangleq \nabla_{\theta_s} \mathcal{L} ( \theta_s, \theta_t, x )
    \end{gathered}
\end{equation}
where $S$ and $\mathcal{L}(.)$ represent the similarity and loss functions. Following \citet{achille2019task2vec}, we use the cosine similarity, indicated by $S_{\cos}$, between the normalized gradients to compute the similarity as follows
\begin{equation*}
    S(g, g^\prime) = S_{\cos} \left( \frac{g}{\vert g \vert + \vert g ^\prime \vert}, \frac{g^\prime}{\vert g \vert + \vert g^\prime \vert} \right)
\end{equation*}
Here, $g$ and $g^\prime$ are shorthands for $g(x, t)$ and $g(x^\prime, t^\prime)$, respectively, and $\vert \cdot \vert$ computes element-wise absolute values. Note that \eqref{eq:task_sim} does not constrain that the inputs must be the same for all tasks and may be applied to tasks not sharing the same input.

We use a pre-trained backbone ($\theta_s$) with task-specific heads ($\theta_t$'s) to compute the similarity. The task-specific heads are randomly initialized, so we fine-tune them on the training examples. The backbone remains frozen during this fine-tuning. For our experiments, the expectation, $\mathbb{E}$, in \eqref{eq:task_sim} is replaced by the empirical mean. As our task similarity is defined as the expectation of example-level similarity, we ensure that no model parameters are disabled in the forward pass due to dropout layers.

\subsection{Task-Grouped LoRA (TGLoRA)}
\label{sec:tglora}
To implement the progressive task-specific architecture, we create a LoRA-based \citep{hu2022lora} layer: Task-Grouped LoRA (TGLoRA). It consists of multiple low-rank modules, one for each task group. It takes as many inputs as the number of task groups and produces the same number of outputs. Let $W \in \mathbb{R}^{d \times k}$ represent a pre-trained weight matrix and $A_\pi \in \mathbb{R}^{r_\pi \times k}$ and $B_\pi \in \mathbb{R}^{d \times r_\pi}$ represent low-rank matrices with rank $r_\pi$ for task group $\pi$. The input corresponding to task group $\pi$, $x_\pi \in \mathbb{R}^k$, is processed as 
\begin{equation*}
    y_\pi = W x_\pi + \gamma_{r, \pi} B_\pi A_\pi x_\pi
\end{equation*}
where $\gamma_{r, \pi} = \frac{\alpha_\pi}{r_\pi}$ represents the scaling factor defined in \citet{hu2022lora}. Figure~\ref{fig:tglora_layer} shows this layer for two task groups. Note that a TGLoRA layer with one task group reduces to a LoRA layer.

\begin{figure}
    \centering


    
    \begin{tikzpicture}[font={\fontfamily{phv}\selectfont\tiny}, text=darkgray!90, draw=darkgray!90]
    \tikzset{
        roundbox/.style={rounded corners=0.1cm},
        shadowbox/.style={blur shadow={shadow blur steps=8}}
    }

    \node [roundbox, shadowbox, fill=BurntOrange!10, minimum height=2.25cm, minimum width=4cm] (TGLORA) {};

    \node [shadowbox, draw=Black, fill=White, text width=0.8cm, text centered, minimum height=0.8cm, minimum width=0.8cm] (PRETRAINED) at ($(TGLORA) - (1.3cm, -0.1cm)$) {\scalebox{0.65}{Pre-Trained} \scalebox{0.65}{$d \times k$}};

    \node [draw=OliveGreen, fill=OliveGreen!10, trapezium, trapezium angle=110, minimum height=0.38cm, anchor=north, blur shadow={shadow blur steps=5}] (LORA1TOP) at ($(PRETRAINED.north) + (1.2cm, 0)$) {\scalebox{0.65}{$d \times r_{1}$}};
    
    \node [draw=OliveGreen, fill=OliveGreen!10, trapezium, trapezium angle=70, minimum height=0.38cm, anchor=south, blur shadow={shadow blur steps=5}] (LORA1BOTTOM) at ($(PRETRAINED.south) + (1.2cm, 0)$) {\scalebox{0.65}{$r_{1} \times k$}};

    \node [draw=CadetBlue, fill=CadetBlue!20, trapezium, trapezium angle=110, minimum height=0.38cm, anchor=north, blur shadow={shadow blur steps=5}] (LORA2TOP) at ($(PRETRAINED.north) + (2.5cm, 0)$) {\scalebox{0.65}{$d \times r_{2}$}};
    
    \node [draw=CadetBlue, fill=CadetBlue!20, trapezium, trapezium angle=70, minimum height=0.38cm, anchor=south, blur shadow={shadow blur steps=5}] (LORA2BOTTOM) at ($(PRETRAINED.south) + (2.5cm, 0)$) {\scalebox{0.65}{$r_{2} \times k$}};

    \node [above=of LORA1TOP, above=1cm] (TASK1) {$T_1$, $T_2$};
    
    \node [above=of LORA2TOP, above=1cm] (TASK2) {$T_3$};
    
    \node[circle, inner sep=0.04cm, draw, fill=none, minimum height=0.1cm, below=of LORA1BOTTOM, below=0.45cm] (ADD1) {\scalebox{0.7}{+}};
    
    \node[circle, inner sep=0.04cm, draw, fill=none, minimum height=0.1cm, below=of LORA2BOTTOM, below=0.45cm] (ADD2) {\scalebox{0.7}{+}};
    
    \node [below=of ADD1, below=0.5cm] (OUT1) {$T_1$, $T_2$};
    
    \node [below=of ADD2, below=0.5cm] (OUT2) {$T_3$};
    
    \draw [OliveGreen, -Stealth] (TASK1.south) -- (LORA1TOP.north);
    
    \draw [CadetBlue, -Stealth] (TASK2.south) -- (LORA2TOP.north);
    
    \draw [OliveGreen, -Stealth] (TASK1.south) -- ++(0, -0.55cm) -- ($(PRETRAINED.north) + (-0.2cm, 0.45cm)$) -- ($(PRETRAINED.north) + (-0.2cm, 0)$);
    
    \draw [CadetBlue, -Stealth] (TASK2.south) -- ++(0, -0.75cm) --  ($(PRETRAINED.north) + (0.2, 0.25cm)$) -- ($(PRETRAINED.north) + (0.2cm, 0)$);
    
    \draw [OliveGreen, -Stealth] (LORA1BOTTOM.south) -- (ADD1.north);
    
    \draw [CadetBlue, -Stealth] (LORA2BOTTOM.south) -- (ADD2.north);
    
    \draw [OliveGreen, -Stealth] ($(PRETRAINED.south) + (-0.2cm, 0)$) -- ($(PRETRAINED.south) + (-0.2cm, -0.58cm)$) -- (ADD1.west);
    
    \draw [CadetBlue, -Stealth] ($(PRETRAINED.south) + (0.2cm, 0)$) -- ($(PRETRAINED.south) + (0.2cm, -0.2cm)$) -- ($(PRETRAINED.south) + (1.5cm, -0.2cm)$) -- ($(PRETRAINED.south) + (1.5cm, -0.58cm)$) -- (ADD2.west);
    
    \draw [OliveGreen, -Stealth] (ADD1.south) -- (OUT1.north);
    
    \draw [CadetBlue, -Stealth] (ADD2.south) -- (OUT2.north);
    
    \node [draw=OliveGreen!10, fill=OliveGreen!20, minimum width=2cm, below=of TGLORA.south west, anchor=west, yshift=-0.2cm] (TG1LEGEND) {\scalebox{0.8}{Task Group 1: $T_1$, $T_2$}};
    
    \node [draw=CadetBlue!20, fill=CadetBlue!20, below=of TGLORA.south east, anchor=east, yshift=-0.2cm] {\scalebox{0.8}{Task Group 2: $T_3$}};
\end{tikzpicture}
    

    \caption{TGLoRA Layer with two task groups. $T_1$, $T_2$, and $T_3$ represent three tasks.}
    \label{fig:tglora_layer}
\end{figure}

\subsection{Computing Task Groups}
\label{sec:task_group_creation}
Creating task groups is a set partitioning problem. Formally, it can be defined as partitioning $\mathcal{T}$ into a partition $P = \left\lbrace \pi_1 \dots \pi_M \right\rbrace$ such that

\begin{itemize}
    \item $\bigcup_{i = 1}^{M} \pi_i = \mathcal{T}$; and
    \item $\pi_i \cap \pi_j = \emptyset \;\forall\; 1 \le i, j \le M ~\text{and}~ i \neq j$
\end{itemize}
We assign a score to every partition $P$, based on the tasks in each group $\pi_i$, which we define momentarily. Intuitively, this score should be higher when similar tasks are grouped and lower when conflicting tasks are assigned to the same task group. This score is determined by evaluating the affinity of each task with other tasks within its respective group. Our objective is to find the partition that maximizes the score.

First, we assign a score to each task $t$ in a task group $\pi_i$ as
\begin{equation*}
    \mathcal{S}_{t, \pi_i} = \begin{cases}
        0 & \vert \pi_i \vert = 1 \\
        \frac{\sum_{t^\prime \in \pi_i, t^\prime \neq t} \operatorname{sim}(t, t^\prime)}{\vert \pi_i \vert - 1} & \text{otherwise}
    \end{cases}
\end{equation*}
where $\operatorname{sim}(t, t^\prime)$ is the task similarity defined in \eqref{eq:task_sim}. Subsequently, the score for partition $P$ is defined as
\begin{equation}\label{eq:partition_score}
    \mathcal{S}_P = \sum_{i = 1}^{M} \sum_{t \in \pi_i} \mathcal{S}_{t, \pi_i}
\end{equation}

To convert a pre-trained model to a branched network, task groups are computed in reverse. We first compute the task groups for the layer just before the decoders. If this layer contains task-specific modules with no sharing, we partition $\mathcal{T}$ into subsets of size one and move to the preceding layer. The branch-and-bound algorithm \citep{fifty2021efficiently} is used to find the partition that maximizes $\mathcal{S}_P$. For other layers, we either keep the same task groups as the succeeding layer or merge them to achieve the required number of groups. Algorithm~\ref{alg:task_group_merging} is used for task group merging.

\begin{algorithm}[t]
\caption{Task Group Merging.}\label{alg:task_group_merging}
\begin{algorithmic}[1]
\Require Tasks $\mathcal{T} = \left\lbrace 1 \dots T \right\rbrace$, a set of task groups $P = \left\lbrace\pi_1 \dots \pi_M\right\rbrace$, task similarities $\left\lbrace S_{t,t^\prime} \vert t, t^\prime \in \mathcal{T} \right\rbrace$, $N < M$

\State $Q \gets P$ \Comment{Partition after merging}

\Repeat
\State $Q^\prime \gets \emptyset$ \Comment{Best partition after merging two groups}
\State $s \gets 0$ \Comment{Score for the best partition}
\ForAll{$\pi_i, \pi_j \in Q$}
\State $P^\prime \gets \left(P \setminus \left\lbrace \pi_i, \pi_j \right\rbrace\right) \cup \left\lbrace \pi_i \cup \pi_j \right\rbrace$
\State $s^\prime \gets \mathcal{S}_{P^\prime}$ \Comment{Defined in \eqref{eq:partition_score}}
\If{$s^\prime > s$}
\State $s \gets s^{\prime}$
\State $Q^\prime \gets P^\prime$
\EndIf
\EndFor
\State $Q \gets Q^{\prime}$
\Until{$\vert Q \vert = N$}
\State \Return $Q$
\end{algorithmic}
\end{algorithm}

\section{EXPERIMENTS}
\label{sec:experiments}
\subsection{Datasets}
Following previous works in the MTL literature \citep{vandenhende2020mti, xu2018pad, ye2022inverted, liu2022polyhistor, agiza2024mtlora}, we evaluate our approach on dense prediction tasks. We use the PASCAL \citep{everingham2010pascal} and \nyud{} \citep{silberman2012indoor} datasets.

For PASCAL, PASCAL-Context split \citep{chen2014detect} is used. It has four dense prediction tasks: (1) semantic segmentation, (2) saliency detection, (3) surface normal estimation, and (4) human part segmentation. The dataset consists of training and validation splits, with 4,998 and 5,105 images, respectively.

We consider three tasks in the \nyud{} dataset: (1) semantic segmentation, (2) depth estimation, and (3) surface normal estimation. The dataset contains 795 training and 654 validation examples.

\subsection{Evaluation Metrics}
The performance on semantic segmentation, saliency detection, and human part segmentation is evaluated using mean intersection-over-union (mIoU). Surface normal estimation and depth estimation are evaluated using the root mean square error (rmse). Following previous works on MTL \citep{vandenhende2021multi, liu2022polyhistor, agiza2024mtlora}, we use the number of trainable parameters and the average per-task difference in performance compared to single-task full fine-tuning ($\Delta m$) to evaluate multi-task models. $\Delta m$ is defined as
\begin{equation*}
    \Delta m = \frac{1}{T} \sum_{t = 1}^{T} \left( -1 \right)^{l_t} \frac{M_{t} - M_{st, t}}{M_{st, t}}
\end{equation*}
where $M_t$ and $M_{st, t}$ represent the multi-task and single-task models' performances on task $t$, respectively. $l_t$ is $0$ if a higher score means better performance for task $t$ and is $1$ otherwise.

\subsection{Setup}
\label{sec:training_details}
\paragraph{Base Model.}
For PASCAL, we replicate the model architecture from \citet{agiza2024mtlora} for a fair comparison with the existing methods. Specifically, we use the Swin Transformer \citep{liu2021swin} and attach task-specific decoders for each task. These decoders are similar to those used in HR-Net \citep{wang2020deep}, consisting of linear and bilinear upsampling layers. We use the Swin-Tiny variant pre-trained on ImageNet-1k. We use the same architecture for \nyud{}. We also experiment with Pyramid Vision Transformer \citep[PVT;][]{wang2021pyramid} for PASCAL and use the PVT-Small variant pre-trained on ImageNet-1k.

Furthermore, we employ multi-scale task-specific feature sharing used in MTLoRA, which merges features from different stages for a comprehensive feature representation. See \citet{agiza2024mtlora} for more details on multi-scale task-specific feature sharing.

\paragraph{Optimization.}
The models are trained to minimize the multi-task loss
\begin{equation}
    L_{\text{MTL}} = \sum_{i = 1}^{T} w_t \times L_t
\end{equation}
where $w_t$ and $L_t$ are the task weight and loss for task $t$. Task weights are the same as in previous works \citep{liu2022polyhistor, agiza2024mtlora}. We use per-pixel cross-entropy loss for semantic and human part segmentation, L1 loss for surface normal and depth estimation, and balanced cross-entropy loss for saliency detection.

\begin{figure}[t]
    \centering
    \begin{subfigure}{0.45\linewidth}
        \centering
        \begin{tikzpicture}[
    font={\fontfamily{phv}\selectfont\tiny},
    text=darkgray!90,
    draw=darkgray!90,
    level distance=1cm,
    level 1/.style={sibling distance=1.5cm},
    level 2/.style={sibling distance=0.8cm},
    tnode/.style={circle, draw, minimum size=0.1cm, inner sep=0.15cm},
]
    \node[tnode] (A) {}
        child { node[tnode] (B) {}
            child { node[tnode] (D) {} 
                child { node[tnode] (G) {} }
                child { node[tnode] (H) {} }
            }
        }
        child { node[tnode] (C) {}
            child { node[tnode] (E) {}
                child { node[tnode] (I) {} }
            }
            child { node[tnode] (F) {} 
                child { node[tnode] (J) {} }
            }
        };

    \node [below=of G, below=0.1cm] {\rotatebox{90}{semseg}};
    \node [below=of H, below=0.1cm] {\rotatebox{90}{h\_parts}};
    \node [below=of I, below=0.1cm] {\rotatebox{90}{sal}};
    \node [below=of J, below=0.1cm] {\rotatebox{90}{normals}};
\end{tikzpicture}
        \caption{PASCAL}
    \end{subfigure}
    \begin{subfigure}{0.45\linewidth}
        \centering
        \begin{tikzpicture}[
    font={\fontfamily{phv}\selectfont\tiny},
    text=darkgray!90,
    draw=darkgray!90,
    level distance=1cm,
    level 1/.style={sibling distance=2cm},
    level 2/.style={sibling distance=1.25cm},
    level 3/.style={sibling distance=0.75cm},
    tnode/.style={circle, draw, minimum size=0.1cm, inner sep=0.15cm},
]
    \node[tnode] (A) {}
        child { node[tnode] (B) {}
            child { node[tnode] (C) {}
                child { node[tnode] (E) {} }
                child { node[tnode] (F) {} }
            }
            child { node[tnode] (D) {} 
                child { node[tnode] (G) {} }
            }
        };

    \node [below=of E, below=0.1cm] {\rotatebox{90}{semseg}};
    \node [below=of F, below=0.1cm] {\rotatebox{90}{depth}};
    \node [below=of G, below=0.1cm] {\rotatebox{90}{normals}};

\end{tikzpicture}
        \caption{\nyud{}}
    \end{subfigure}
    \caption{Task groups computed for PASCAL (with Swin and PVT) and \nyud{} (with Swin).}
    \label{fig:best_groups_similarity}
\end{figure}
\begin{table*}[t]
    \caption{Comparison of progressive task-specific adaptation (indicated by TGLoRA) with the existing methods on PASCAL using a Swin backbone. $\uparrow$ and $\downarrow$ signify that higher and lower values are better, respectively. Results marked with \textdagger ~are from \citet{agiza2024mtlora}.}
    \label{tab:pascal_results}
    \centering
    \scalebox{0.95}{
    \begin{small}
    \begin{tabular}{l|r@{}lr@{}lr@{}lr@{}l|r|r}
    \toprule
    \multirow{2}{*}{\textbf{Method}} & \multicolumn{2}{c}{\textbf{SemSeg}} & \multicolumn{2}{c}{\textbf{Human Parts}} & \multicolumn{2}{c}{\textbf{Saliency}} & \multicolumn{2}{c|}{\textbf{Normals}} & $\Delta m$ & \textbf{Trainable} \\
    
    {} & \multicolumn{2}{c}{(mIoU $\uparrow$)} & \multicolumn{2}{c}{(mIoU $\uparrow$)} & \multicolumn{2}{c}{(mIoU $\uparrow$)} & \multicolumn{2}{c|}{(rmse $\downarrow$)} & (\% $\uparrow$) & \textbf{Params} (M $\downarrow$) \\
    \midrule
    Single Task -- Full Fine-Tuning\textsuperscript{\textdagger} & 67.21 & {} & 61.93 & {}  & 62.35 & {}  & 17.97 & {}  & 0 & 112.62 \\
    MTL -- Decoders Only\textsuperscript{\textdagger} & 65.09 & {} & 53.48 & {} & 57.46 & {}  & 20.69 & {}  & $-$9.95 & 1.94 \\
    MTL -- Full Fine-Tuning\textsuperscript{\textdagger} & 67.56 & {}  & 60.24 & {}  & 65.21 & {}  & 16.64 & {}  & $+$2.23 & 30.06 \\
    \midrule
    Hyperformer\textsuperscript{\textdagger} & 71.43 & {}  & 60.73 & {}  & 65.54 & {}  & 17.77 & {}  & $+$2.64 & 72.77 \\
    Polyhistor\textsuperscript{\textdagger} & 70.87 & {}  & 59.15 & {}  & 65.54 & {}  & 17.77 & {}  & $+$2.34 & 8.96 \\
    MTLoRA\textsuperscript{\textdagger} ($r = 64$) & 67.90 & {}  & 59.84 & {}  & 65.40 & {}  & 16.60 & {}  & $+$2.55 & 8.34 \\
    MTL -- LoRA ({\footnotesize \attnmlponly{}}) & 67.57 & {\tiny $\pm$0.19} & 59.09 & {\tiny $\pm$0.10} & 65.18 & {\tiny $\pm$0.07} & 17.08 & {\tiny $\pm$0.03} & $+$1.36 & 5.29 \\
    Single Task -- LoRA ({\footnotesize \attnmlponly{}}) & 70.80 & {\tiny $\pm$0.08} & 58.73 & {\tiny $\pm$0.07} & 66.05 & {\tiny $\pm$0.04} & 16.65 & {\tiny $\pm$0.06} & $+$3.36 & 5.29 \\

    \midrule

    TGLoRA ({\footnotesize \attnmlponly{} + \other{}}) & 70.53 & {\tiny $\pm$0.08} & 60.96 & {\tiny $\pm$0.12} & 66.12 & {\tiny $\pm$0.04} & 16.43 & {\tiny $\pm$0.03} & $+$4.50 & 6.89 \\
    TGLoRA ({\footnotesize \attnmlponly{}}) & 70.39 & {\tiny $\pm$0.29} & 60.58 & {\tiny $\pm$0.10} & 65.94 & {\tiny $\pm$0.01} & 16.61 & {\tiny $\pm$0.01} & $+$3.97 & 5.29 \\
    TGLoRA ({\footnotesize \attnonly{}}) & 70.27 & {\tiny $\pm$0.09} & 59.36 & {\tiny $\pm$0.04} & 65.38 & {\tiny $\pm$0.15} & 17.09 & {\tiny $\pm$0.04} & $+$2.54 & 3.20 \\

    \bottomrule
    \end{tabular}
    \end{small}
    }
\end{table*}
\begin{table*}[t]
    \caption{Comparison of progressive task-specific adaptation (indicated by TGLoRA) with the existing methods on \nyud{} using a Swin backbone. $\uparrow$ and $\downarrow$ signify that higher and lower values are better, respectively.}
    \label{tab:nyud_results}
    \centering
    \scalebox{0.95}{
    \begin{small}
    \begin{tabular}{l|r@{}lr@{}lr@{}l|r|r}
    \toprule
    \multirow{2}{*}{\textbf{Method}} & \multicolumn{2}{c}{\textbf{SemSeg}} & \multicolumn{2}{c}{\textbf{Normals}} & \multicolumn{2}{c|}{\textbf{Depth}} & $\Delta m$ & \textbf{Trainable} \\
    {} & \multicolumn{2}{c}{(mIoU $\uparrow$)} & \multicolumn{2}{c}{(rmse $\downarrow$)} & \multicolumn{2}{c|}{(rmse $\downarrow$)} & (\% $\uparrow$) & \textbf{Params} (M $\downarrow$) \\
    \midrule
    Single Task -- Full Fine-Tuning & 41.85 & {\tiny $\pm$0.37} & 24.01 & {\tiny $\pm$0.05} & 0.6322 & {\tiny $\pm$0.0014} & 0 & 84.04 \\
    MTL -- Full Fine-Tuning & 41.17 & {\tiny $\pm$0.44} & 24.75 & {\tiny $\pm$0.02} & 0.6217 & {\tiny $\pm$0.0019} & $-$1.01 & 29.00 \\
    MTL -- Decoder Only & 35.97 & {\tiny $\pm$0.02} & 32.63 & {\tiny $\pm$0.02} & 0.8008 & {\tiny $\pm$0.0012} & $-$25.54 & 1.48 \\
    \midrule
    MTLoRA ($r = 64$) & 41.52 & {\tiny $\pm$0.27} & 24.99 & {\tiny $\pm$0.04} & 0.6212 & {\tiny $\pm$0.0014} & $-$1.04 & 7.81 \\
    MTL -- LoRA ({\footnotesize \attnmlponly{}}) & 41.05 & {\tiny $\pm$0.28} & 25.01 & {\tiny $\pm$0.03} & 0.6231 & {\tiny $\pm$0.0019} & $-$1.54 & 5.93 \\
    Single Task -- LoRA ({\footnotesize \attnmlponly{}}) & 41.86 & {\tiny $\pm$0.17} & 24.57 & {\tiny $\pm$0.03} & 0.6295 & {\tiny $\pm$0.0022} & $-$0.63 & 5.93 \\
    \midrule
    TGLoRA ({\footnotesize \attnmlponly{} + \other{}}) & 41.84 & {\tiny $\pm$0.29} & 24.38 & {\tiny $\pm$0.06} & 0.6177 & {\tiny $\pm$0.0030} & $+$0.24 & 7.53 \\
    TGLoRA ({\footnotesize \attnmlponly{}}) & 41.89 & {\tiny $\pm$0.24} & 24.45 & {\tiny $\pm$0.03} & 0.6236 & {\tiny $\pm$0.0039} & $-$0.12 & 5.93 \\
    TGLoRA ({\footnotesize \attnonly{}}) & 41.92 & {\tiny $\pm$0.28} & 25.27 & {\tiny $\pm$0.04} & 0.6319 & {\tiny $\pm$0.0039} & $-$1.68 & 3.15 \\
    \bottomrule
    \end{tabular}
    \end{small}
    }
\end{table*}

\paragraph{Trainable Layers.}
We conduct our experiments by adding TGLoRA modules to the attention (indicated by \attn{}) and MLP (indicated by \mlp{}) layers. We determine the rank of the low-rank matrices in each task group of a TGLoRA layer based on the number of tasks in that group. Using a combined rank $r$ for a TGLoRA layer, we allocate it proportionally across different task groups. This strategy ensures better control of the trainable parameter count across different configurations. Additionally, similar to MTLoRA, we explore unfreezing the patch embedding, patch merging, layer normalization, and position bias layers (indicated by \other{}).

\paragraph{Training Details.}
We repeat each experiment three times with different random seeds and report the average scores. Further details on the training setup are provided in Section~\ref{appendix:hyperparameters} of the supplementary material.

\paragraph{Baselines.}
We consider the following baselines: (1) Polyhistor \citep{liu2022polyhistor}, (2) MTLoRA \citep{agiza2024mtlora}, and (3) Hyperformer \citep{mahabadi2021parameter, liu2022polyhistor}.  Additionally, we include the following two baselines:
\vspace{-0.5em}
\begin{itemize}
    \item \textit{Single Task -- LoRA:} We fine-tune single-task models by adding LoRA modules to various layers in the model.

    \item \textit{MTL -- LoRA:} We also fine-tune multi-task models by adding LoRA modules to various layers in the model.
\end{itemize}

We scale the ranks of adapter modules in these two baselines to match the number of trainable parameters with TGLoRA. These baselines represent individual single-task and shared multi-task adaptations.

\begin{table*}[t]
    \caption{Comparison of progressive task-specific adaptation (indicated by TGLoRA) with the existing methods on PASCAL using a PVT backbone. $\uparrow$ and $\downarrow$ signify that higher and lower values are better, respectively. Results marked with \textdagger ~are from \citet{agiza2024mtlora}.}
    \label{tab:pascal_pvt_results}
    \centering
    \scalebox{0.95}{
    \begin{small}
    \begin{tabular}{l|r@{}lr@{}lr@{}lr@{}l|r|r}
    \toprule
    \multirow{2}{*}{\textbf{Method}} & \multicolumn{2}{c}{\textbf{SemSeg}} & \multicolumn{2}{c}{\textbf{Human Parts}} & \multicolumn{2}{c}{\textbf{Saliency}} & \multicolumn{2}{c|}{\textbf{Normals}} & $\Delta m$ & \textbf{Trainable} \\
    {} & \multicolumn{2}{c}{(mIoU $\uparrow$)} & \multicolumn{2}{c}{(mIoU $\uparrow$)} & \multicolumn{2}{c}{(mIoU $\uparrow$)} & \multicolumn{2}{c|}{(rmse $\downarrow$)} & (\% $\uparrow$) & \textbf{Params} (M $\downarrow$) \\
    \midrule
    Single Task -- Full Fine-Tuning\textsuperscript{\textdagger} & 68.81 & {} & 61.27 & {} & 62.67 & {} & 17.55 & {} & 0 & 97.51 \\
    MTL -- Decoders Only\textsuperscript{\textdagger} & 64.86 & {} & 51.18 & {} & 61.54 & {} & 19.55 & {} & $-$8.85 & 2.11 \\
    \midrule
    Hyperformer\textsuperscript{\textdagger} & 70.81 & {} & 57.76 & {} & 65.49 & {} & 17.75 & {} & +0.14 & 16.14 \\
    Polyhistor\textsuperscript{\textdagger} & 71.00 & {} & 57.52 & {} & 65.83 & {} & 17.83 & {} & +0.13 & 7.32 \\
    MTLoRA\textsuperscript{\textdagger} ($r = 64$) & 69.74 & {} & 58.08 & {} & 65.62 & {} & 17.35 & {} & +1.20 & 8.69 \\

    MTLoRA (Reproduced, {\footnotesize \attnonly{} + \textsc{sr}}) & 70.32 & {\tiny $\pm$0.41} & 59.06 & {\tiny $\pm$0.10} & 66.10 & {\tiny $\pm$0.07} & 16.99 & {\tiny $\pm$0.02} & +1.81 & 8.69 \\
    MTLoRA (Reproduced, {\footnotesize \attnonly{}}) & 70.34 & {\tiny $\pm$0.40} & 58.83 & {\tiny $\pm$0.13} & 65.89 & {\tiny $\pm$0.08} & 17.15 & {\tiny $\pm$0.03} & +1.42 & 4.40 \\

    \midrule

    TGLoRA ({\footnotesize \attnonly{}}) & 72.90 & {\tiny $\pm$0.17} & 60.15 & {\tiny $\pm$0.13} & 66.76 & {\tiny $\pm$0.02} & 16.91 & {\tiny $\pm$0.03} & +3.57 & 3.85 \\
    
    \bottomrule
    \end{tabular}
    \end{small}
    }
\end{table*}

\subsection{Results}
\label{sec:main_results}

\paragraph{Task Groups.}
For both datasets, we share adapter modules across tasks in the first stage, keep the last stage task-specific, and reduce the number of task groups as we move away from the decoders. Figure~\ref{fig:best_groups_similarity} shows the task groups computed by our method for PASCAL and \nyud{}. Notably, both the Swin and PVT backbones result in the same task groups for PASCAL. More details on task group creation and the effect of different computational budgets are presented in Sections~\ref{appendix:task_group_creation} and \ref{appendix:compute_budget} of the supplementary material.

\paragraph{MTL Performance.}
Tables~\ref{tab:pascal_results} and \ref{tab:nyud_results} show the per-task performance, the overall MTL performance ($\Delta m$), and the number of trainable parameters for PASCAL and \nyud{}, respectively, with a Swin backbone. For PASCAL, our approach achieves an absolute improvement of more than 2\% in overall MTL performance over MTLoRA and Polyhistor and uses significantly fewer trainable parameters. Furthermore, it surpasses Hyperformer by a significant margin, which requires considerably more trainable parameters. Similar to \citet{agiza2024mtlora}, we also observe improved performance on unfreezing additional layers compared to fine-tuning only the newly added TGLoRA modules. As a final point, adding TGLoRA only to the attention layers is very effective and achieves comparable performance to the existing methods while requiring less than half the trainable parameters. For \nyud{}, our approach outperforms MTLoRA by more than 1\% with fewer trainable parameters. Moreover, it achieves better overall MTL performance than the existing methods while keeping additional layers frozen during fine-tuning, leading to a significantly better trade-off between the number of trainable parameters and performance.

For both datasets, the progressive task-specific architecture outperforms the single-task (Single Task -- LoRA) and multi-task models with a shared backbone (MTL -- LoRA) for the same parameter budget. This behavior remains consistent when TGLoRA is added only to the attention layer and when additional layers are unfrozen during training. See Section~\ref{appendix:more_results} in the supplementary material for the results with these configurations.

\begin{table}[t]
    \setlength{\tabcolsep}{3pt}
    \centering
    \caption{Giga multiply-accumulate operations for individual single-task, shared multi-task, and progressive task-specific multi-task adaptations using a Swin backbone.}
    \label{tab:flops}
    \resizebox{0.97\linewidth}{!}{
    \begin{small}
    \begin{tabular}{lcccc}
    \toprule
    \multirow{2}{*}{\bf Dataset} & {\bf \#} & {\bf Individual} & {\bf Shared} & {\bf Progressive} \\
    {} & {\bf Tasks} & {\bf Single-Task} & {\bf Multi-Task} & {\bf Task-Specific} \\
    \midrule
    \multirow{2}{*}{PASCAL} & 1 & 18.49 & 18.49 & 18.49 \\
    {} & 4 & 73.96 & 21.47 & 49.95 \\
    \midrule
    \multirow{2}{*}{\nyud{}} & 1 & 18.56 & 18.56 & 18.56 \\
    {} & 3 & 55.41 & 20.53 & 34.65 \\
    \bottomrule
    \end{tabular}
    \end{small}
    }
\end{table}

We also experiment with PVT to ensure that our method generalizes to different backbones. We follow the training setup from \citet{agiza2024mtlora} and fine-tune PVT-Small on the PASCAL dataset. For both MTLoRA and TGLoRA, adapter modules are added to the attention layer (indicated by \attn{}). Additionally, for MTLoRA, the sequence reduction layers are unfrozen (indicated by \attn{} + \textsc{sr}). Table~\ref{tab:pascal_pvt_results} shows the performance with PVT. These results demonstrate that TGLoRA outperforms the existing methods with fewer trainable parameters. They further show that unfreezing the sequence reduction layers in MTLoRA contributes little to the performance and significantly increases the trainable parameters.

Lastly, in addition to adding adapter modules to different layers, we control the number of trainable parameters by varying the ranks of individual adapter modules. We observe a similar pattern as reported in Table~\ref{tab:pascal_results} for different values of ranks. See Section~\ref{appendix:more_results} in the supplementary material for results.

\paragraph{Inference Cost.}
Table~\ref{tab:flops} shows the number of giga multiply-accumulate operations for our approach compared to individual single-task and shared multi-task adaptations. These results show that the inference cost for progressive task-specific adaptations lies between the two extremes.

\begin{table}[t]
    \caption{Performance of progressive task-specific adaptation with TGLoRA when less similar tasks are in the same task group. $\Delta m$ is computed based on the single-task performance from Table~\ref{tab:pascal_results} for PASCAL and Table~\ref{tab:nyud_results} for \nyud{}. These experiments use a Swin backbone.}
    \label{tab:pascal_nyud_ablations}
    \centering
    \resizebox{\linewidth}{!}{
    \begin{small}
    \begin{tabular}{llrr}
    \toprule
    \multirow{2}{*}{\textbf{Dataset}} & \multirow{2}{*}{\textbf{Method}} & $\Delta m$ & \textbf{Trainable} \\
    {} & {} & (\% $\uparrow$) & \textbf{Params} (M $\downarrow$) \\
    \midrule
    \multirow{3}{*}{PASCAL} & \all{} & +3.71 & 6.89 \\
    {} & \attnmlponly{} & +3.14 & 5.29 \\
    {} & \attnonly{} & +1.95 & 3.20 \\

    \midrule

    \multirow{3}{*}{\nyud{}} & \all & $-$1.11 & 7.53 \\
    {} & \attnmlponly{} & $-$0.90 & 5.93 \\
    {} & \attnonly{} & $-$2.30 & 3.15 \\
    
    \bottomrule
    \end{tabular}
    \end{small}
    }

\end{table}

\subsection{Sub-Optimal Task Groups}
Next, we examine the impact of grouping less similar tasks, as per the task similarities computed in the previous section (shown in Figure~\ref{fig:best_groups_similarity}), while maintaining the number of task groups. For PASCAL, we swap saliency and human part segmentation in the second and third stages of Swin. Similarly, we swap depth and surface normal estimation tasks in the third stage for \nyud{}. The MTL performance under these configurations is illustrated in Table~\ref{tab:pascal_nyud_ablations}. These results demonstrate that grouping less similar tasks adversely impacts the model's performance, providing strong evidence in favor of our proposed method of computing task similarity and forming task groups.

\section{LIMITATIONS}
In our experiments, the number of task groups in different stages of Swin and PVT is selected manually. Although we report results for several branching configurations, exploring a broader range could yield deeper insights and strengthen our approach. An important future direction is to develop automated methods for selecting the optimal number of task groups in different layers. Another limitation lies in the task grouping algorithm itself. We use a branch-and-bound-style procedure to compute task groups in the final stage. While effective for a small number of tasks, this approach becomes intractable as the number of tasks grows. An approximate variant of the algorithm could make our method more scalable to large-scale MTL scenarios.

\section{CONCLUSION}
\label{sec:conclusion}
In this work, we introduced progressive task-specific multi-task adaptation, positioned between individual single-task and shared multi-task adaptations. To implement it, we developed a new LoRA-based layer named TGLoRA and a gradient-based metric to measure task similarity. We adapted the Swin Transformer and Pyramid Vision Transformer for dense prediction tasks. Our experiments demonstrate that progressive task-specific adaptation outperforms both individual single-task and shared multi-task adaptations, achieving significantly greater relative improvement over single-task models while using fewer trainable parameters than the current state-of-the-art approaches. Ablation studies further reveal that grouping less similar tasks adversely affects multi-task performance, highlighting the effectiveness of our method. Finally, our approach offers flexible control over the trade-off between computational budget and performance.

\subsubsection*{Acknowledgements}
This work was supported in part by the Illinois Campus Cluster Program and the Delta Advanced Computing and Data Resource. Delta is supported by the National Science Foundation (award OAC 2005572) and the State of Illinois and is a joint effort of the University of Illinois Urbana-Champaign and its National Center for Supercomputing Applications. We also thank the anonymous reviewers for their feedback.

\bibliographystyle{apalike}
\bibliography{main}

\section*{Checklist}

\begin{enumerate}

  \item For all models and algorithms presented, check if you include:
  \begin{enumerate}
    \item A clear description of the mathematical setting, assumptions, algorithm, and/or model. [Yes, Section~\ref{sec:proposed_approach}]
    \item An analysis of the properties and complexity (time, space, sample size) of any algorithm. [Yes, Sections~\ref{sec:proposed_approach} and \ref{sec:experiments}]
    \item (Optional) Anonymized source code, with specification of all dependencies, including external libraries. [No]
  \end{enumerate}

  \item For any theoretical claim, check if you include:
  \begin{enumerate}
    \item Statements of the full set of assumptions of all theoretical results. [Not Applicable]
    \item Complete proofs of all theoretical results. [Not Applicable]
    \item Clear explanations of any assumptions. [Not Applicable]     
  \end{enumerate}

  \item For all figures and tables that present empirical results, check if you include:
  \begin{enumerate}
    \item The code, data, and instructions needed to reproduce the main experimental results (either in the supplemental material or as a URL). [\url{https://github.com/neerajgangwar/progressive-task-specific-adaptation}]
    \item All the training details (e.g., data splits, hyperparameters, how they were chosen). [Yes, Section~\ref{appendix:hyperparameters} in the supplementary material]
    \item A clear definition of the specific measure or statistics and error bars (e.g., with respect to the random seed after running experiments multiple times). [Yes]
    \item A description of the computing infrastructure used. (e.g., type of GPUs, internal cluster, or cloud provider). [Yes]
  \end{enumerate}

  \item If you are using existing assets (e.g., code, data, models) or curating/releasing new assets, check if you include:
  \begin{enumerate}
    \item Citations of the creator if your work uses existing assets. [Yes]
    \item The license information of the assets, if applicable. [Not applicable]
    \item New assets either in the supplemental material or as a URL, if applicable. [Not Applicable]
    \item Information about consent from data providers/curators. [Not Applicable]
    \item Discussion of sensible content if applicable, e.g., personally identifiable information or offensive content. [Not Applicable]
  \end{enumerate}

  \item If you used crowdsourcing or conducted research with human subjects, check if you include:
  \begin{enumerate}
    \item The full text of instructions given to participants and screenshots. [Not Applicable]
    \item Descriptions of potential participant risks, with links to Institutional Review Board (IRB) approvals if applicable. [Not Applicable]
    \item The estimated hourly wage paid to participants and the total amount spent on participant compensation. [Not Applicable]
  \end{enumerate}

\end{enumerate}

\clearpage
\appendix
\onecolumn

\thispagestyle{empty}
\aistatstitle{Parameter-Efficient Multi-Task Learning via Progressive\\Task-Specific Adaptation: Supplementary Materials}

\section{TRAINING HYPERPARAMETERS}
\label{appendix:hyperparameters}
\paragraph{PASCAL.}
We replicate the hyperparameters from \citet{agiza2024mtlora} for fine-tuning the models on the PASCAL dataset. Specifically, we use the AdamW optimizer \citep{Loshchilov2017DecoupledWD} with a batch size of 32, a learning rate of 3.125 $\times \text{10}^{-\text{5}}$, and a weight decay of 0.05. The models are fine-tuned for 300 epochs, with evaluations every 20 epochs. We use a linear warmup for the first 20 epochs, followed by a cosine annealing learning rate scheduler. For the LoRA modules, a dropout rate of 0.05 and a scaling ($\gamma_{r, \pi}$) of 4 are used. We also apply the following data augmentations: \texttt{RandomHorizontalFlip}, \texttt{RandomScale} ranging from 0.75 to 1.25, and \texttt{RandomRotate} ranging from -20 to 20. We use combined ranks ($r$) of 64, 32, and 16 for TGLoRA. For Swin, we use a fixed rank of four for all task-specific modules in the last stage. Although we keep the rank constant in the last stage of Swin, it could also be scaled according to $r$, which would increase the number of trainable parameters.  Our experiments use A100 GPUs (40GB VRAM).

\paragraph{\nyud{}.}
We use a batch size of 6 and a learning rate of $\text{10}^{-\text{4}}$ for \nyud{}. We use a combined rank of 64 for TGLoRA. We also apply the following data augmentations: \texttt{RandomHorizontalFlip} and \texttt{RandomScale} with possible scales of 1, 1.2, and 1.5. All other hyperparameters remain the same as those in PASCAL. Our experiments use A100 GPUs (40GB VRAM).

\section{TASK GROUPS}
\label{appendix:task_group_creation}
We determine the number of task groups in each TGLoRA layer according to a predefined computational budget, such as parameter count or floating-point operations per second (FLOPS). Fixing the computational budget fixes the branching in the architecture. For simplicity, we do not change the number of task groups in the TGLoRA layers within a stage of Swin or PVT.

As discussed in Section~\ref{sec:task_similarity}, fine-tuning the decoders is necessary for computing task similarity. We examine the impact of fine-tuning them with a limited subset of training examples versus the full training dataset. Our experiments show that both approaches produce identical task groups. Furthermore, repeating the fine-tuning process three times with different random seeds confirms the consistency of the results. These results confirm that this fine-tuning step introduces minimal overhead in the training pipeline.

\begin{table*}[p]
    \caption{Additional results on the PASCAL dataset using a Swin backbone. The number of trainable parameters is controlled by varying the rank of the LoRA modules. $\Delta m$ is computed based on the single-task performance from Table~\ref{tab:pascal_results}.}
    \label{tab:pascal_tglora_varying_ranks}
    \centering
    \scalebox{0.95}{
    \begin{small}
    \begin{tabular}{l|r@{}lr@{}lr@{}lr@{}l|r|r}
    \toprule
    \multirow{2}{*}{\textbf{Trainable Layers}} & \multicolumn{2}{c}{\textbf{SemSeg}} & \multicolumn{2}{c}{\textbf{Human Parts}} & \multicolumn{2}{c}{\textbf{Saliency}} & \multicolumn{2}{c|}{\textbf{Normals}} & $\Delta m$ & \textbf{Trainable} \\
    {} & \multicolumn{2}{c}{(mIoU $\uparrow$)} & \multicolumn{2}{c}{(mIoU $\uparrow$)} & \multicolumn{2}{c}{(mIoU $\uparrow$)} & \multicolumn{2}{c|}{(rmse $\downarrow$)} & (\% $\uparrow$) & \textbf{Params} (M $\downarrow$) \\

    \midrule
    \multicolumn{11}{c}{\it TGLoRA} \\
    \midrule
    \multirow{2}{*}{{\footnotesize \all{}}} & 70.66 & {\tiny $\pm$0.04} & 60.48 & {\tiny $\pm$0.11} & 65.56 & {\tiny $\pm$0.12} & 16.65 & {\tiny $\pm$0.04} & $+$3.82 & 5.41 \\
    {} & 70.66 & {\tiny $\pm$0.07} & 59.76 & {\tiny $\pm$0.08} & 65.15 & {\tiny $\pm$0.02} & 16.79 & {\tiny $\pm$0.08} & $+$3.17 & 4.68 \\
    \midrule
    \multirow{2}{*}{{\footnotesize \attnmlponly{}}} & 70.40 & {\tiny $\pm$0.07} & 60.01 & {\tiny $\pm$0.03} & 65.34 & {\tiny $\pm$0.12} & 16.90 & {\tiny $\pm$0.03} & $+$3.10 & 3.81 \\
    {} & 70.34 & {\tiny $\pm$0.26} & 59.17 & {\tiny $\pm$0.06} & 64.84 & {\tiny $\pm$0.08} & 17.13 & {\tiny $\pm$0.03} & $+$2.21 & 3.08 \\
    \midrule
    \multirow{2}{*}{{\footnotesize \attnonly{}}} & 70.33 & {\tiny $\pm$0.09} & 58.90 & {\tiny $\pm$0.10} & 64.72 & {\tiny $\pm$0.11} & 17.45 & {\tiny $\pm$0.05} & $+$1.61 & 2.65 \\
    {} & 70.27 & {\tiny $\pm$0.11} & 58.37 & {\tiny $\pm$0.03} & 64.09 & {\tiny $\pm$0.20} & 17.69 & {\tiny $\pm$0.06} & $+$0.79 & 2.37 \\
    \midrule
    \multicolumn{11}{c}{\it MTL -- LoRA} \\
    \midrule
    \multirow{3}{*}{{\footnotesize \all{}}} & 67.62 & {\tiny $\pm$0.27} & 59.22 & {\tiny $\pm$0.06} & 65.28 & {\tiny $\pm$0.09} & 16.86 & {\tiny $\pm$0.05} & $+$1.78 & 6.89 \\
    {} & 68.07 & {\tiny $\pm$0.08} & 59.07 & {\tiny $\pm$0.05} & 64.96 & {\tiny $\pm$0.05} & 17.11 & {\tiny $\pm$0.05} & $+$1.41 & 5.41 \\
    {} & 68.52 & {\tiny $\pm$0.08} & 58.78 & {\tiny $\pm$0.06} & 64.29 & {\tiny $\pm$0.03} & 17.30 & {\tiny $\pm$0.01} & $+$0.93 & 4.68 \\
    \midrule
    \multirow{2}{*}{{\footnotesize \attnmlponly{}}} & 68.02 & {\tiny $\pm$0.04} & 58.78 & {\tiny $\pm$0.05} & 64.77 & {\tiny $\pm$0.03} & 17.33 & {\tiny $\pm$0.01} & $+$0.89 & 3.81 \\
    {} & 68.35 & {\tiny $\pm$0.14} & 58.30 & {\tiny $\pm$0.07} & 63.99 & {\tiny $\pm$0.09} & 17.62 & {\tiny $\pm$0.03} & $+$0.10 & 3.08 \\
    \midrule
    \multirow{3}{*}{{\footnotesize \attnonly{}}} & 68.25 & {\tiny $\pm$0.11} & 58.42 & {\tiny $\pm$0.07} & 64.38 & {\tiny $\pm$0.09} & 17.57 & {\tiny $\pm$0.04} & $+$0.34 & 3.20 \\
    {} & 68.52 & {\tiny $\pm$0.06} & 58.11 & {\tiny $\pm$0.10} & 63.73 & {\tiny $\pm$0.05} & 17.92 & {\tiny $\pm$0.02} & $-$0.43 & 2.65 \\
    {} & 68.62 & {\tiny $\pm$0.13} & 57.65 & {\tiny $\pm$0.16} & 63.23 & {\tiny $\pm$0.12} & 18.20 & {\tiny $\pm$0.03} & $-$1.17 & 2.37 \\
    \midrule
    \multicolumn{11}{c}{\it Single Task -- LoRA} \\
    \midrule
    \multirow{3}{*}{\footnotesize \all{}} & 71.16 & {\tiny $\pm$0.06} & 59.42 & {\tiny $\pm$0.09} & 66.21 & {\tiny $\pm$0.10} & 16.41 & {\tiny $\pm$0.06} & $+$4.17 & 11.69 \\
    {} & 71.32 & {\tiny $\pm$0.04} & 59.05 & {\tiny $\pm$0.08} & 65.78 & {\tiny $\pm$0.11} & 16.58 & {\tiny $\pm$0.03} & $+$3.68 & 10.21 \\
    {} & 71.37 & {\tiny $\pm$0.08} & 58.70 & {\tiny $\pm$0.03} & 65.35 & {\tiny $\pm$0.03} & 16.67 & {\tiny $\pm$0.04} & $+$3.25 & 9.48 \\
    \midrule
    \multirow{3}{*}{\footnotesize \attnmlponly{}} & 70.83 & {\tiny $\pm$0.13} & 58.15 & {\tiny $\pm$0.04} & 65.54 & {\tiny $\pm$0.08} & 16.95 & {\tiny $\pm$0.10} & $+$2.52 & 3.81 \\
    {} & 70.82 & {\tiny $\pm$0.08} & 57.70 & {\tiny $\pm$0.04} & 64.97 & {\tiny $\pm$0.05} & 17.14 & {\tiny $\pm$0.05} & $+$1.84 & 3.08 \\
    \midrule
    \multirow{3}{*}{\footnotesize \attnonly{}} & 70.65 & {\tiny $\pm$0.05} & 57.78 & {\tiny $\pm$0.03} & 65.20 & {\tiny $\pm$0.10} & 17.25 & {\tiny $\pm$0.03} & $+$1.75 & 3.20 \\
    {} & 70.63 & {\tiny $\pm$0.11} & 57.47 & {\tiny $\pm$0.09} & 64.75 & {\tiny $\pm$0.15} & 17.48 & {\tiny $\pm$0.05} & $+$1.12 & 2.65 \\
    {} & 70.52 & {\tiny $\pm$0.12} & 56.99 & {\tiny $\pm$0.03} & 64.11 & {\tiny $\pm$0.10} & 17.74 & {\tiny $\pm$0.02} & $+$0.26 & 2.37 \\

    \bottomrule
    \end{tabular}
    \end{small}
    }
\end{table*}
\begin{table*}[p]
    \caption{Additional results on the \nyud{} dataset using a Swin backbone. $\Delta m$ is computed based on the single-task performance from Table~\ref{tab:nyud_results}.}
    \label{tab:nyud_additional_results}
    \centering
    \begin{small}
    \begin{tabular}{l|r@{}lr@{}lr@{}l|r|r}
    \toprule
    \multirow{2}{*}{Trainable Layers} & \multicolumn{2}{c}{\textbf{SemSeg}} & \multicolumn{2}{c}{\textbf{Normals}} & \multicolumn{2}{c|}{\textbf{Depth}} & $\Delta m$ & \textbf{Trainable} \\
    {} & \multicolumn{2}{c}{(mIoU $\uparrow$)} & \multicolumn{2}{c}{(rmse $\downarrow$)} & \multicolumn{2}{c|}{(rmse $\downarrow$)} & (\% $\uparrow$) & \textbf{Params} (M $\downarrow$) \\
    \midrule
    \multicolumn{9}{c}{\it MTL -- LoRA} \\
    \midrule
    {\footnotesize \all{}} & 41.03 & {\tiny $\pm$0.33} & 25.01 & {\tiny $\pm$0.05} & 0.6216 & {\tiny $\pm$0.0040} & $-$1.48 & 7.53 \\
    {\footnotesize \attnonly{}} & 41.92 & {\tiny $\pm$0.43} & 25.81 & {\tiny $\pm$0.05} & 0.6309 & {\tiny $\pm$0.0028} & $-$2.37 & 3.15 \\
    \midrule
    \multicolumn{9}{c}{\it Single Task -- LoRA} \\
    \midrule
    {\footnotesize \all{}} & 42.10 & {\tiny $\pm$0.15} & 24.49 & {\tiny $\pm$0.04} & 0.6310 & {\tiny $\pm$0.0047} & $-$0.40 & 10.73 \\
    \midrule
    {\footnotesize \attnonly{}} & 41.69 & {\tiny $\pm$0.21} & 25.34 & {\tiny $\pm$0.02} & 0.6376 & {\tiny $\pm$0.0048} & $-$2.26 & 3.15 \\
    \bottomrule
    \end{tabular}
    \end{small}
\end{table*}

\section{ADDITIONAL EXPERIMENTS}
\subsection{Computational Budget vs Performance}
\label{appendix:compute_budget}
The tree structure in our approach offers a trade-off between model performance and inference cost. For instance, using 6.89M parameters for PASCAL using a Swin backbone, and assigning one, one, two, and four task groups to the first through fourth stages, respectively, results in $\Delta m = +3.93\%$ with 38.37 GMacs. Similarly, configuring the stages with one, two, two, and four task groups results in $\Delta m = +4.21\%$ and 41.38 GMacs.

\subsection{Detailed Results for PASCAL and \nyud{}}
\label{appendix:more_results}
In this section, we present additional experiments on the PASCAL and \nyud{} datasets. Continuing from Section~\ref{sec:main_results}, Table~\ref{tab:pascal_tglora_varying_ranks} illustrates the performance of TGLoRA for varying trainable parameters. The rank of the low-rank modules in TGLoRA layers controls this number. Ranks for adapter modules in \textit{Single Task -- LoRA} and \textit{MTL -- LoRA} are chosen to keep the number of trainable parameters the same as TGLoRA. The table also shows the performance of \textit{Single Task -- LoRA} and \textit{MTL -- LoRA} when the attention and MLP layers are augmented with TGLoRA and when the patch embedding, patch merging, layer normalization, and position bias layers are unfrozen during fine-tuning. Note that unfreezing additional layers results in more trainable parameters in the single-task models, as these layers are unfrozen in the individual models. Hence, the \attn{} + \mlp{} + \other{} configuration has more trainable parameters than the corresponding TGLoRA or \textit{MTL -- LoRA} counterparts. Our experiments demonstrate that progressive task-specific adaptation outperforms individual single-task and shared multi-task adaptations when different layers are augmented with TGLoRA or when additional layers are unfrozen during fine-tuning.

\end{document}